\documentclass[runningheads]{llncs}
\usepackage[T1]{fontenc}
\usepackage{graphicx}

\usepackage{subfigure} %
\usepackage{multirow} %
\usepackage{multicol}
\usepackage{xcolor}

\usepackage[misc,geometry]{ifsym}

\begin{document}
\title{Investigating Guiding Information for Adaptive Collocation Point Sampling in PINNs}
\titlerunning{Guiding Information for Adaptive Sampling}
\author{Jose Florido\inst{1}\textsuperscript{(\Letter)}\orcidID{0009-0001-7291-5029} \and %
He Wang\inst{2}\orcidID{0000-0002-2281-5679} \and %
Amirul Khan\inst{1}\orcidID{0000-0002-7521-5458} \and %
Peter K. Jimack\inst{1}\orcidID{0000-0001-9463-7595}} %
\authorrunning{J. Florido et al.}
\institute{University of Leeds, Leeds LS2 9JT, UK
\email{\{mn17jilf,A.Khan,P.K.Jimack\}@leeds.ac.uk} \and
University College London, London WC1E 6BT, UK \\
\email{he\_wang@ucl.ac.uk}}
\maketitle              %
\begin{abstract}
Physics-informed neural networks (PINNs) provide a means of obtaining approximate solutions of partial differential equations and systems through the minimisation of an objective function which includes the evaluation of a residual function at a set of collocation points within the domain. The quality of a PINNs solution depends upon numerous parameters, including the number and distribution of these collocation points. In this paper we consider a number of strategies for selecting these points and investigate their impact on the overall accuracy of the method. In particular, we suggest that no single approach is likely to be ``optimal'' but we show how a number of important metrics can have an impact in improving the quality of the results obtained when using a fixed number of residual evaluations. We illustrate these approaches through the use of two benchmark test problems: Burgers' equation and the Allen-Cahn equation.

\end{abstract}
\section{Introduction}
\label{sec:1}
\subsection{Context}
This paper is concerned with the mechanism by which Physics-informed neural networks (PINNs) \cite{lagaris,raissi}, allow us to introduce our prior knowledge of physics into a machine learning (ML) algorithm. PINNs are an ML approach to the solution of systems of partial differential equations (PDEs), where there is limited or noisy ground truth data.
In this context, the PDEs provide a mathematical model of the physics, which is captured in the method through the inclusion of loss terms that are
evaluated by calculating PDE residuals at collocation points throughout the domain. This point-based, meshless approach offers greater flexibility compared to meshed methods in traditional numerical models - but also poses an interesting problem regarding the number and distribution of said collocation points.

The original implementation of PINNs uniformly distributes a fixed number of collocation points randomly throughout the domain. However, in certain problems some areas are intrinsically harder to learn than others - and this purely random approach can lead to slow or inefficient training. Different collocation points contain different amounts of information for learning, which indicates there is an underlying distribution of locations for collocation points that can maximise the learning - which is mostly likely non-uniform and problem dependent. Such a proposition suggests that adjusting the distribution of points, by biasing towards features of interest, may be beneficial in terms of efficiency. It may also be necessary for practical reasons to minimise the risk of the network getting trapped in local minima during training~\cite{wang}. However, manually choosing this distribution of collocation points can be an arduous task and requires {\em a priori} knowledge of the solution, or at least of which areas of the domain will be ``most important''. The alternative is to increase the number of points used globally which, whilst potentially effective, adds significantly to the computational cost.

To find the optimal or a good distribution of the collocation points, there are two common approaches to explicitly or implicitly approximate this distribution. The first approach is to have fixed collocation points but weight them differently, i.e. \textit{adaptive weighting}. Examples of this are \cite{li}, \cite{mcclenny}'s Self-Adaptive PINNs, a concept extended by \cite{zhang}'s DASA point weighing, and more recently \cite{song}'s Loss-Attentional PINNs. The residual-based attention scheme in \cite{anagnostopoulos} works similarly to the above, weighting the influence of specific collocation points in the domain to ensure key collocation points aren't overlooked.

The other approach is to refine the locations of collocation points, i.e. \textit{adaptive resampling}. \cite{lu} first presents this type of adaptive refinement based on residual information. Other sampling approaches include \cite{nabian,hanna,gao}. \cite{wu}'s formulation for resampling is a more general version of some of the previous approaches, and systematically compares adaptive collocation resampling methods to fixed approaches. Also noteworthy is \cite{subramanian}'s implementation of a cosine-annealing strategy for restarting training from a uniform distribution when optimisation stalls. Whilst the above focus on the collocation points where the PDEs are evaluated, \cite{lau} also considers optimising the selection of experimental points (where the data is available for supervised learning problems) at the same time.

From the sampling point of view, existing methods essentially seek the ideal distribution of collocation points, via either explicit parameterisation \cite{lu}, or implicit approximation \cite{nabian}. The information exploited up to date is mostly focused on the loss function \cite{nabian,gao,wu}, e.g. information such as local PDE residual at collocation points, or gradients of the loss term. Deviating from existing literature, we investigate into a different category of information that proves to be useful in parametrising the ideal collocation points. This information is the geometric information of the estimated solution, e.g. their spatial and temporal derivatives. The intuition behind is the solution geometry reveals intrinsic information about the PDE. For instance, stiffness regions are harder to learn \cite{song}.
By introducing the solution geometry, our aim is to shed a deeper light into the complexity behind seeking an ``ideal'' distribution, and into the interplay between the effectiveness of different point sampling methods, the complexity of the problem and the computational cost.

\subsection{PINNs}\label{sec:1.2}
The process of training a PINN is similar to that for a regular NN: once a topology and a set of weights is defined we seek to find values of these weights that minimise a prescribed loss function. As discussed below, we will consider a fully unsupervised learning task for which the loss function
is obtained from the sum of all of the residuals of the PDEs at all of the selected collocation points. The significance of the number and location of the collocation points is substantial therefore, since they directly impact on the loss function which is to be minimised. The crux of adaptive resampling is that a fixed number of points are moved based upon some criteria (that we investigate), with a view to this modified loss function being a better objective for the overall minimisation . Previous investigations of this approach have redistributed the collocation points based upon a probability density function (PDF) that was formed using the values of the residuals at the current points or the derivatives of the loss function with respect to the location of the collocation points. In this work, we consider alternative criteria based upon mixed derivatives of the estimated solution and of the residual.

The loss function \( \mathcal{L}(\theta) \) of a PINN can be typically characterised as the sum of the individual losses due to data fitting (\( \mathcal{L}_{data} \)), enforcing the governing PDE (\( \mathcal{L}_{PDE} \)), and weakly enforcing boundary conditions (\( \mathcal{L}_{BCs} \)) as follows:
\[
\mathcal{L}(\theta) = w_{data}\mathcal{L}(\theta)_{data} + w_{PDE}\mathcal{L}(\theta)_{PDE} + w_{BC}\mathcal{L}(\theta)_{BCs}
\]
Here, the relative impact of each loss term can be controlled through the weight, $w$, whilst $\theta$ represents the trainable network parameters. For the one-dimensionsal time-dependent problems that we look at, we will consider a spatial dimension $x$, a temporal dimension $t$ and solve for the dependent variable $u(x,t)$. The PDE loss term is obtained by evaluating the PDE residual ($f$, say) for every collocation point in the set of selected collocation points $\mathrm{x} \in \mathcal{T}$:
\[
\mathcal{L}_{PDE} = \frac{1}{|\mathcal{T}|}\sum_{\mathrm{x}\in \mathcal{T}}|f(\mathrm{x};u;u_x,u_t,u_{xx};\theta)|^2
\]
It is important to distribute these points throughout the domain, however, evidence suggests that it may also be beneficial to identify and place a greater proportion of points in particular regions of importance, so that the loss term especially includes these areas. This can ensure that optimising the cost function improves how well the PDE is satisfied around these locations, and therefore hopefully improves the solution in those areas.

For forward problems, PINNs can be trained without the need for data, and boundary conditions can be enforced in a `hard' manner. This is the approach that we take here in order to focus exclusively on the PDE residual contribution to the loss. We strongly impose the boundary conditions by applying an output transformation to the results of the neural network and we treat the PINN as an unsupervised problem with no input data. As a result, $\mathcal{L}_{PDE}$ becomes the only term to optimise in our loss function.

Throughout this paper, for consistency with other studies in the literature~\cite{wu}, a simple feedforward NN is used consisting of 3 intermediate layers of 64 nodes each, with a tanh activation function after each layer. The training consists of 15,000 initial steps using the ADAM optimiser (with learning rate of 0.001), followed by 1000 steps using L-BFGS \cite{wu} before beginning the resampling process for adaptive methods. For these methods, the points are then redistributed at this stage and then the training continues with 1000 steps of ADAM and 1000 steps of L-BFGS, repeating until the number of resamples specified are completed.

\section{Adaptive resampling} \label{sec:2}

\subsection{Probability density functions}
As mentioned in the introduction, for many problems PINNs will give more accurate results when the collocation points are distributed following an appropriate, problem-dependent distribution. The process of rearranging the collocation points can be automated by using information gathered during the training process. This approach has been shown to work well in the literature using the local residual as a guiding metric \cite{lu,wu}; but this paper also proposes alternatives based on the spatial and temporal derivatives of both the residuals %
and estimates of the solution.
These proposed alternatives are assessed against both uniform, fixed distributions and existing adaptive redistribution methods, and shown to be beneficial in many cases.

\begin{figure}
    \centering
    \includegraphics[width=\linewidth]{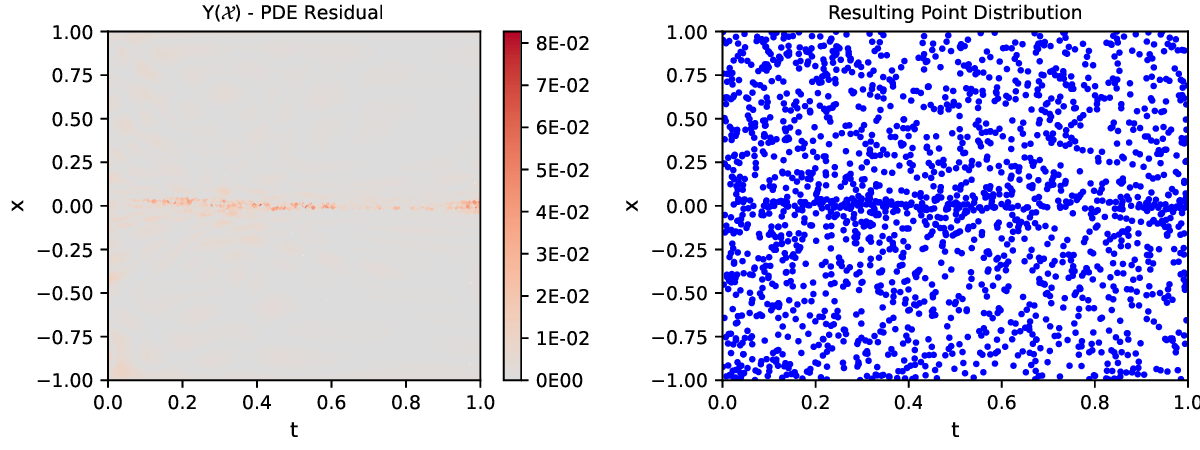}
    \caption{$Y(\mathcal{X})=$ PDE Residual shown at every point in $\mathcal{X}$ (left); and the corresponding points selected for $\mathcal{T}$ (right).}
    \label{fig:example_distribution}
\end{figure}

The method that we use to resample the points is consistent with \cite{wu} for comparison, which chooses the next set of collocation points $\mathcal{T}$ from a fine grid of random points $\mathcal{X}$. All points in $\mathcal{X}$ are assigned a probability ($P(\mathcal{X})$), and the prescribed number of collocation points are chosen according to the normalised PDF $\hat{P}(\mathcal{X})$. These are obtained from equations \ref{eq:kc} and \ref{eq:kc2}:

\begin{equation}
P(\mathcal{X}) = \frac{{Y(\mathcal{X})}^k}{\overline{Y(\mathcal{X})^k}} + c \label{eq:kc} 
\end{equation}
\begin{equation}
\hat{P}(\mathcal{X}) = \frac{P(\mathcal{X})}{\| P(\mathcal{X}) \|_1} \label{eq:kc2}
\end{equation}
Here $Y(\mathcal{X})$ is the vector containing the magnitude of the information source chosen (e.g. the current residual or its derivatives, or some derivatives of the current estimate of the solution), $\overline{Y(\mathcal{X})^k}$ is the mean value across all points in $\mathcal{X}$ and $k, c$ are constant hyper-parameters that affect the resampling behaviour. 
A visualisation of the distribution of points obtained using the PDE residual as $Y(\mathcal{X})$ for solving the Burgers' Equation is shown in figure \ref{fig:example_distribution}.

As a baseline, we also compare the performance of the adaptive methods that we consider against two fixed distributions: a uniform random distribution and an alternative uniform, pseudo-random, distribution made with a Hammersley sampling algorithm \cite{hammersley}. For these fixed distributions, instead of resampling after a given number of steps, the training is continued until an equivalent amount of training steps are carried out.

We consider two cases for the adaptive method for which $Y$ is based upon the current local residuals: one where the initial distribution of points before any resamples is randomly generated ($PDE, R$), and the other where the initial distribution follows Hammersley sampling ($PDE, H$).
We also consider the use of the current solution estimates as an information source, defining $Y$ based upon the mixed (spatial and temporal) second derivative of the current solution. We will refer to this as the local geometric curvature of $u$, and denote this case as ``$U_{xt}$'' in the subsequent text. We also consider a version of $Y$ that is based upon the mixed second derivative of the local residual values, which will subsequently be denoted as ``$PDE_{xt}$''.

\subsection {Problem Definition}
\label{sec:2.2}
\begin{figure}[htbp]
    \centering
    \includegraphics[width=0.75\linewidth]{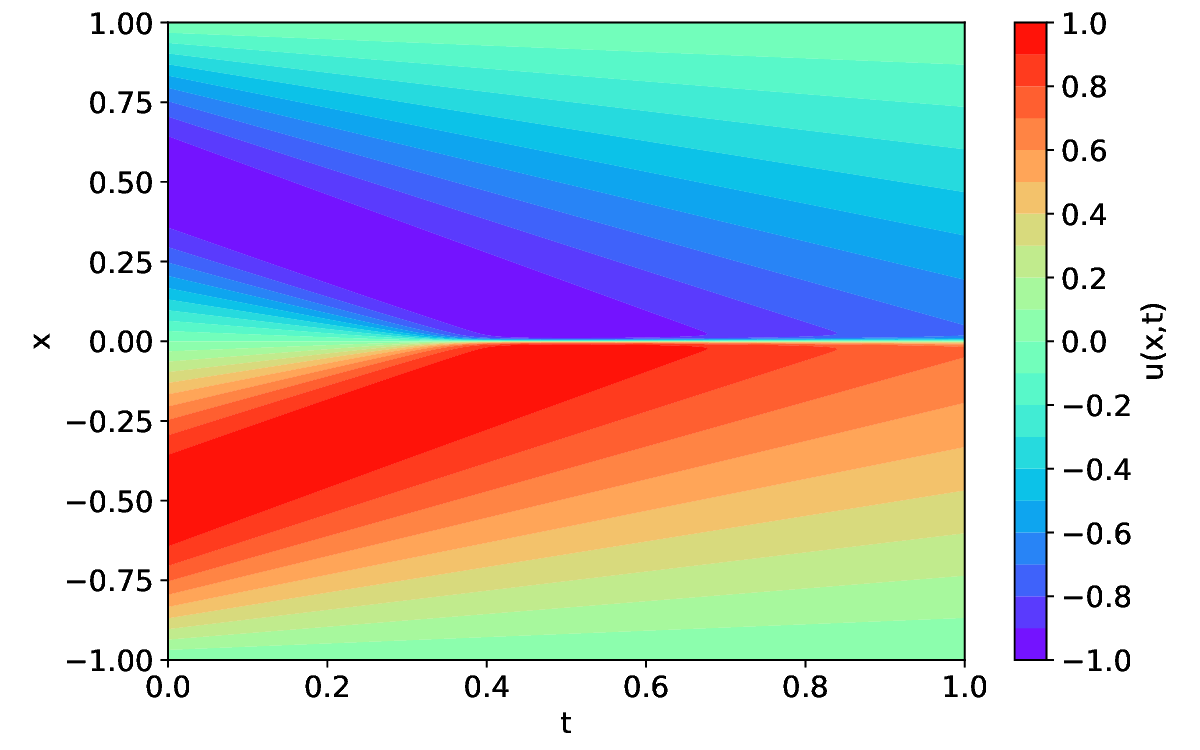}
    \caption{Burgers' Equation contour plot of example solution with $\nu = \frac{1}{100\pi}$}
    \label{fig:sample-sol}
\end{figure}

The main benchmark that we initially use in order to assess the different sampling strategies that we consider is the 1D Burgers' Equation, given by:
\begin{equation}
uu_x + u_t  = \nu \; u_{xx},\quad x \in [-1, 1], \quad t \in [0, 1],
\end{equation}
where the magnitude of $\nu$ determines the relative effect of diffusion.
Dirichlet boundary conditions are assigned:
\[
    u(-1, t) = u(1, t) = 0,
\]
as is the following initial condition:
\[
    u(x, 0) = -\sin(\pi x).
\]
Note that, since we solve this problem on a space-time domain (see figure \ref{fig:sample-sol}) the initial condition and the boundary conditions are treated identically by the PINN. To illustrate the main features of the analytical solution to this problem a colour map of $u$ is plotted in figure \ref{fig:sample-sol}. Note that the smooth initial solution (the left boundary of the domain shown) steepens to a very sharp front by $t=0.25$ and this remains an important feature of the solution for all subsequent times.

As discussed in the previous section the boundary and initial conditions may be imposed either weakly or strongly when using PINNs. Here we opt for the latter by applying an output transformation to the $u$ that is output by the network:
\[
    output = -\sin(\pi*x)+(1-x^2)ut
\]

For this case; the PDE residual when used as an information source is therefore given by:

\begin{equation}
PDE = uu_x + u_t  - \nu \; u_{xx}
\end{equation}
Similarly,
\begin{equation}
U_{xt}= \frac{\partial^2}{\partial x \partial t}u(x,t)
\end{equation}
and
\begin{equation}
PDE_{xt} = \frac{\partial ^2}{\partial x \partial t} (uu_x + u_t  - \nu \; u_{xx})
\end{equation}

To measure and compare the accuracy of the different methods considered an \(L^2\) relative error metric is used at the end of training. This compares the prediction $u$  to ground truth $u_{gt}$ as follows:
\[L^2 = %
\frac{\sqrt{\sum (u(i)-u_{gt}(i))^2}}{\sqrt{\sum u_{gt}(i)^2}} \;. \]
Note that the sums in this last expression are over every point $i$ in a very fine background grid which contains many more points than the number of collocation points used to evaluate $u$. Furthermore, due to the stochastic nature of ML training, a minimum of 20 repeats with different seeds were carried out for each case investigated throughout this paper; with the quoted $L^2$ errors always representing the average of these.

\subsection{Results: default case}
\begin{figure}[htbp]
    \centering
    \includegraphics[width=0.7\textwidth]{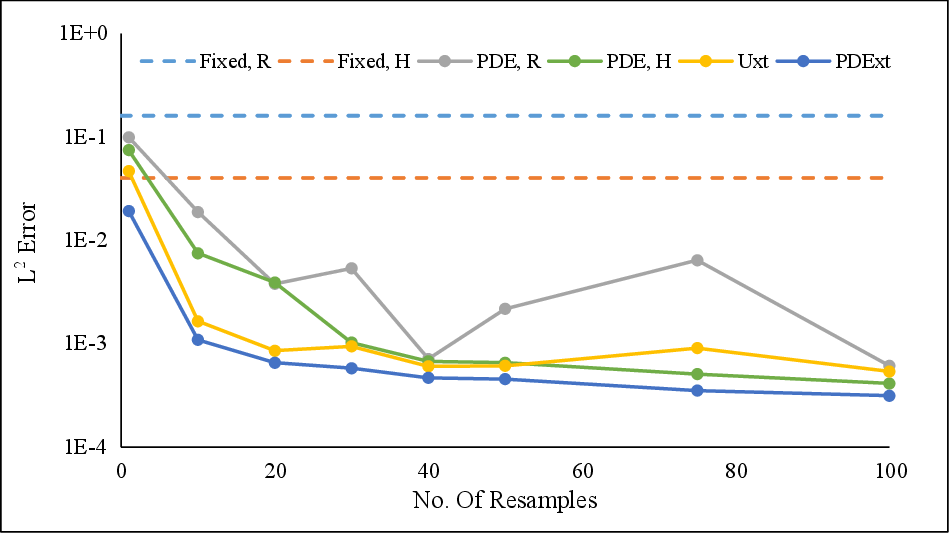}
    \caption{Error against number of resamples solving Burgers' Equation with described default parameters $v=\frac{1}{100\pi}; N=2000$}
    \label{fig:ic1_vs_l}
\end{figure}

For this first test problem we have applied six different solution strategies using $N=2000$ collocation points. In each case we follow the training regime of \cite{wu}, described in section \ref{sec:1.2} (15,000 initial steps of ADAM followed by 1000 steps of L-BFGS and then resampling every 2000 steps (1000 ADAM/1000 L-BFGS). Figure \ref{fig:ic1_vs_l} plots the error against the number of resamples taken for each of the approaches considered, including the two constant point sets based upon a uniform random distribution and a distribution based upon the Hammersley sampling algorithm.

The initial distribution is seen to have an effect in both the fixed and residual-based methods, with the use of the Hammersley sampling algorithm resulting in consistently lower error.
The local residual method with random initialisation ($PDE, R$) exhibited the least robustness and greatest variability, which can be seen from the fact increasing the number of resamples did not consistently lower the average error between runs.
Generally, more resamples does result in improving accuracy, with error descending below 0.1\% for above 30 resamples for most of the adaptive resampling methods.

\begin{figure}[htbp]
    \centering
    \includegraphics[width=0.7\textwidth]{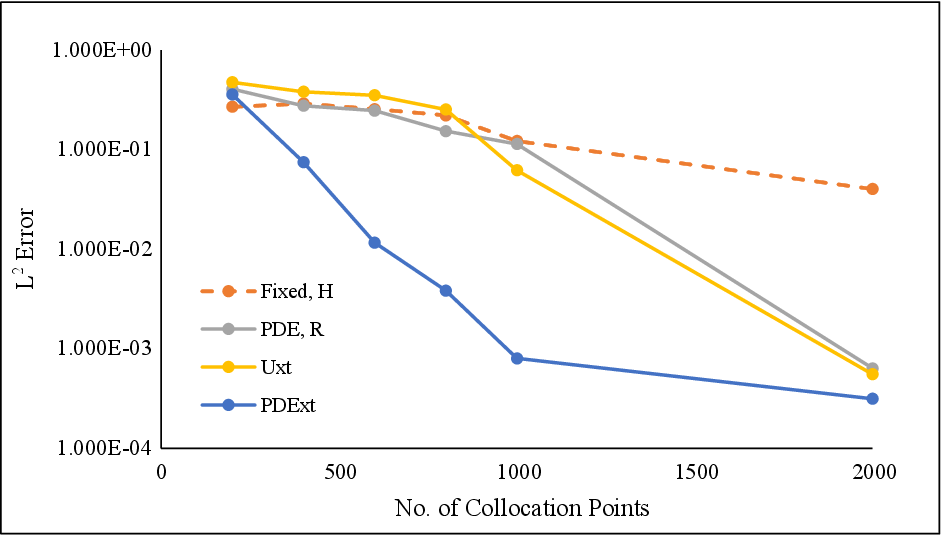}
    \caption{Error against number of collocation points for different resampling methods solving Burgers' Equation with described default parameters $v=\frac{1}{100\pi};$ 100 resamples}
    \label{fig:ic1_vs_n}
\end{figure}

For the same case, we also look at the error achieved using different numbers of collocation points. This was obtained for 100 rounds of resampling and is plotted in figure \ref{fig:ic1_vs_n}.
This figure suggests that the threshold for the number of points required to obtain results of a given accuracy varies from method to method. Even for the most accurate method ($PDE_{xt}$), the error seems to plateau between 0.01\% and 0.1\%, which the other adaptive methods eventually reach at the default 2000 points. Nevertheless, this clearly suggests that the $PDE_{xt}$ approach could allow similar accuracy to other techniques but at a significantly lower cost.

In conclusion, based on this single test case, we observe that the achieved error of a PINN is heavily influenced by the chosen collocation point distribution, echoing what has been seen previously (see Introduction). Furthermore, the adaptive resampling methods easily outperform the fixed distributions, with both of the derivative-based redistribution methods performing quite well (especially when considering the mixed second derivative of the residual). In the next section we explore the extent to which this behaviour generalises by changing multiple aspects of the problem under consideration, including the initial conditions and the magnitude of the diffusion term, and by considering a second benchmark PDE: the Allen-Cahn Equation.

\section{Results: other cases}
\subsection{Alternate Initial Conditions} \label{sec:results-other}
\begin{figure}[htbp]
    \centering
    \includegraphics[width=0.9\textwidth]{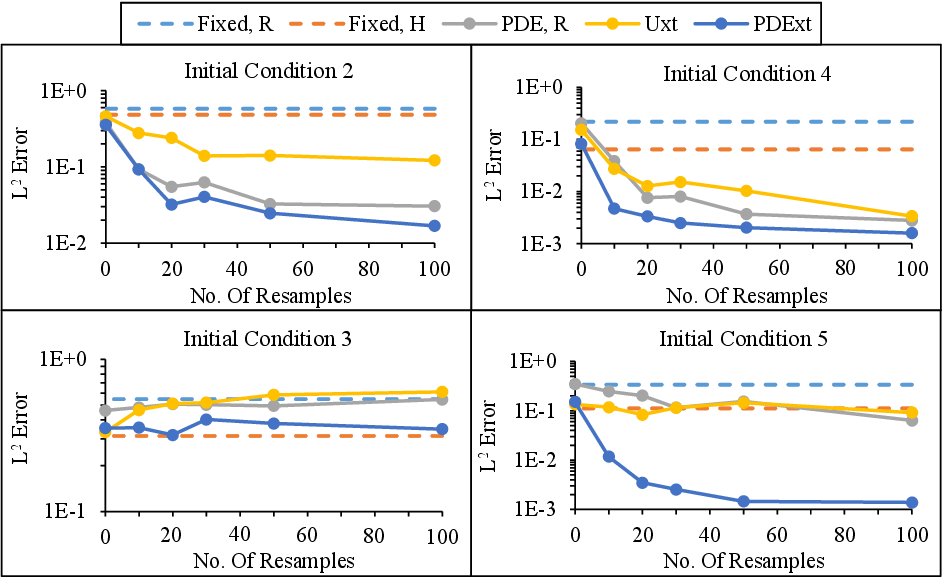}
    \caption{Error versus number of resamples. Initial conditions 2 and 3 are combinations of randomly weighted sin curves of different frequency; with 2 being made up of three low frequency curves and 3 being a higher frequency combination of four curves. Initial conditions 4 and 5 are $sin(2\pi)$ and $1.5sin(\pi)$ respectively. \label{fig:ics}}
\end{figure}

In this example we contrast the same methods as in the previous section to once again solve Burgers' Equation, but this time we vary the initial conditions, and therefore the entire evolution of the solution of the PDE. For each of the four cases considered we compute a high resolution numerical solution to represent the ground truth and use this to assess the error against an increasing number of adaptive resamples of the collocation points. The corresponding results are plotted in figure \ref{fig:ics}.

Inspection of figure \ref{fig:ics} suggests that for three of the four initial conditions (2, 4 and 5), $PDE_{xt}$ clearly outperforms the other methods (as it did for initial condition 1 in the previous section). Not only does it deliver a smaller error after the last resample (the maximum amount of training investigated) but the error curve lies beneath the others at each stage in the adaptive process.  %
This gap is most noticeable in the 5th case, which corresponds to an initial condition with a larger amplitude, thus leading to a faster shock formation and a sharper shock. In this case it far outperforms the next best method.
Furthermore, in each case other than initial condition 3, the adaptive collocation point algorithms all performed significantly better than using fixed collocation points. Unsurprisingly, of the two fixed point approaches Hammersley always outperforms the uniform random distribution.

The obvious outlier in figure  \ref{fig:ics} is initial condition 3. Here all of the methods to quite poorly in terms of reducing the error and least worst result is obtained using fixed collocation points based upon a Hammersley sampling. Further investigation is required to better understand this result, which shows that the PINN performance (or, at least, the ease with which the PINN can be trained) is highly dependent upon the problem being solved.

\subsection{Adjusting PDE parameters} \label{sec:3.4}
\begin{figure}
    \centering
    \includegraphics[width=0.7\textwidth]{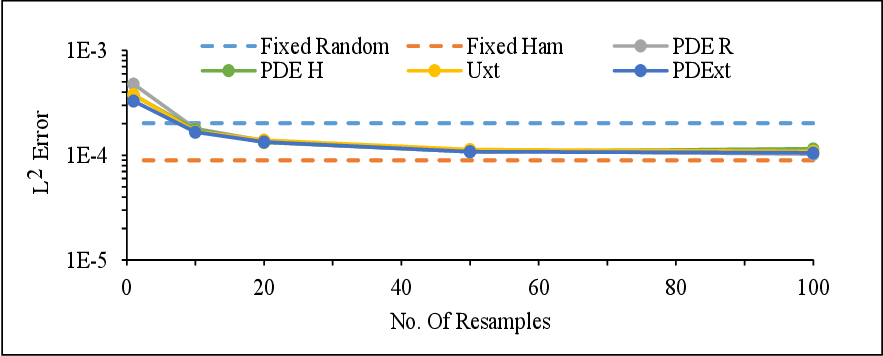}
    \caption{Error against number of resamples, for Burgers' Equation with $\nu = 0.01$}
    \label{fig:visc01}
\end{figure}

\begin{figure}
    \centering
    \includegraphics[width=0.7\textwidth]{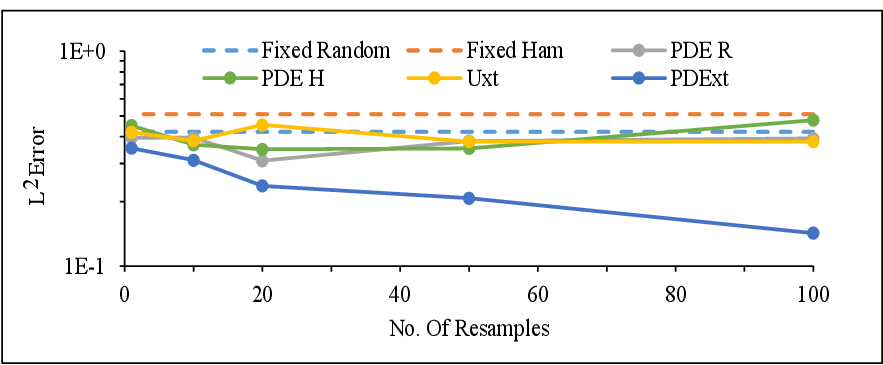}
    \caption{Error against number of resamples, for Burgers' Equation with $\nu = 0.001$}
    \label{fig:visc001}
\end{figure}

For the following example, we investigate adjusting the value of the $\nu$ parameter in the PDE; which affects the magnitude of the diffusion term $\nu \; u_{xx}$. Lower values increase the shock sharpness, increasing the complexity of the problem without fundamentally changing the shape of the solution. Increasing $\nu$ sees the opposite effect. We use values of $\nu$ of $0.01$ and $0.001$, approximately multiplying and dividing by 3 from the default $\nu = \frac{1}{100\pi}$ used in the previous section. The ground truth is again computed using high-resolution numerical simulations and the error when using different information sources to control the adaptive resampling is again compared against the fixed uniform methods: see figures \ref{fig:visc01} and \ref{fig:visc001}.

For the simpler case, $\nu = 0.01$ (figure \ref{fig:visc01}), all adaptive methods behave similarly, delivering a small error after relatively few resamples. Comparing the fixed methods, the Hammerlsey distribution is again consistently better than a random fixed distribution. Furthermore, in this specific example, it again outperforms the adaptive sampling methods. Unlike case 3 of figure  \ref{fig:ics} however, the accuracy of all of these PINNs solutions is high. One possible explanation for the Hammersley approach being the best in this case is that, unlike the resampling methods, it uses the L-BFGS optimisation algorithm for the majority of training steps (as opposed to an equal split between ADAM and L-BFGS for the adaptive methods). 
Overall, It is clear that by making the solution of the problem smoother (through increasing $\nu$) all of the methods do well and therefore use of the simplest approach, based upon a fixed distribution of collocation points, can be appropriate. 

For the more challenging case of $\nu = 0.001$ (figure \ref{fig:visc001}), all methods struggle to converge to an accurate solution using $N=2000$ collocation points. The least worst approach, with an $L^2$ error of 14\%, is again the adaptive method based upon $PDE_{xt}$. Based upon this set of results we can make the opposite proposition: that for complex problems, the choice of adaptive sampling method may not be the limiting factor to accuracy, and an increase to either the number of collocation points or the network complexity may be needed to increase accuracy.

We test this proposition by undertaking a study to look at the impact that changing the number of collocation points has on the solutions obtained in the $\nu = 0.01$ and $\nu = 0.001$ cases. In table \ref{tab:v01} we consider the simpler $\nu=0.01$ case, observing the change in performance when the number of collocation points is halved to $1000$. Conversely, in table \ref{tab:v001}, the more complex case with the sharper shock ($\nu=0.001$), we double the number of collocation points to $N=4000$ and observe whether there are any significant increases in accuracy.
\begin{table}[htbp]
    \centering
    \caption{Effect of changing $N$; $L^2$ Error at 100 resamples for $\nu = 0.01$}
    \label{tab:v01}
    \begin{tabular}{|c|c|c|c|c|c|c|}
    \hline
        $N$ & Fixed, R & Fixed, H & $PDE, R$ & $PDE, H$  & $U_{xt}$ & $PDE_{xt}$ \\\hline
        2000 & 2.019E-04 & 8.95E-05 & 1.024E-04 & 1.152E-04 & 1.089E-04 & 1.055E-04 \\\hline
        1000 & 3.089E-03 & 3.925E-04 & 1.503E-04 & 1.867E-04 & 1.458E-04 & 1.258E-04 \\\hline
        Error Change & 1429.9\% & 338.4\% &  46.8\% & 62.1\% & 33.9\% & 19.3\% \\\hline
    \end{tabular}
\end{table}

In table \ref{tab:v01}, decreasing the number of collocation points from 2000 to 1000 increases the error as expected. However, there is a big discrepancy in how much the different methods are affected. A positive \% signifying increase in error, the fixed methods are substantially less effective with fewer points. The opposite is true for adaptive methods, especially for $PDE_{xt}$ which only sees an increase in error of 19.3\% compared to the 3-fold increase of the fixed Hammersley distribution of points.

\begin{table}[htbp]
    \centering
    \caption{Effect of changing $N$; $L^2$ Error at 100 resamples for $\nu=0.001$}
    \label{tab:v001}
    \begin{tabular}{|c|c|c|c|c|c|c|}
    \hline
        $N$   & Fixed, R & Fixed, H & $PDE, R$  & $PDE, H$  & $U_{xt}$     & $PDE_{xt}$   \\ \hline
        2000 & 4.209E-01    & 5.093E-01 & 3.925E-01 &  4.780E-01        & 3.797E-01 & 1.425E-01 \\ \hline
        4000 & 4.049E-01    & 3.836E-01 & 1.858E-01 & 1.996E-01 & 2.591E-01 & 3.660E-02 \\ \hline
        Error Change & -3.9\% & -32.8\% & -111.2\% & -139.5\% & -46.6\% & -289.3\% \\\hline
        
\end{tabular}
\end{table}

In table \ref{tab:v001}, we are this time looking at the decrease in error as a result of doubling the number of collocation points in the case of $\nu=0.001$. As anticipated, the error does decrease in each column (signified by a negative increase) however this improvement is relatively small in most cases, meaning that the overall errors are still not competitive, even with this increase in resources. The biggest decreases are seen for the $PDE, H$ and especially $PDE_{xt}$, which is the only method to achieve an error under 10\%.

\subsection{An alternative PDE: Allen-Cahn}\label{sec:4}
Having presented several variations of the Burgers' Equation problem, our next step is to consider a different PDE entirely and observe the results. The Allen-Cahn equation is chosen for this due to its complex transient behaviours which are known to be challenging to capture computationally. The problem is defined as follows:
\begin{equation}
\frac{\partial u}{\partial t} = D \frac{\partial^2 u}{\partial x^2} +5(u-u^3), \quad x \in [-1,1], \quad t \in [0,1]
\end{equation}
We take D = 0.001 and use the initial condition
\[
u(x,0) = x^2 \cos(\pi x) \;,
\]
and boundary conditions
\[
u(-1, t) = u(1, t) = -1 \;.
\]

For this problem we again compute a high-resolution numerical solution for the ground truth and consider the $L^2$ error as the number of resamples is increased. Based on the observations of the previous subsections, we undertake these tests for three distinct numbers of collocation points: $N = 500, 1000, 2000$. The same resampling methods as in section \ref{sec:2} are used, using a minimum of 20 repeats per case. The results for $N=500$ are shown below in figure \ref{fig:ac1}.
For this case, the adaptive methods are again consistently superior to using fixed collocation points. Similar to the Burger's equation solutions with a large $\nu = 0.01$, there are no clear advantages between $PDE_{xt}$, and the approaches based upon the local residual (though using $U_{xt}$ is not so good).
\begin{figure}[htbp]\label{fig:ac1}
    \centering
    \includegraphics[width=0.7\linewidth]{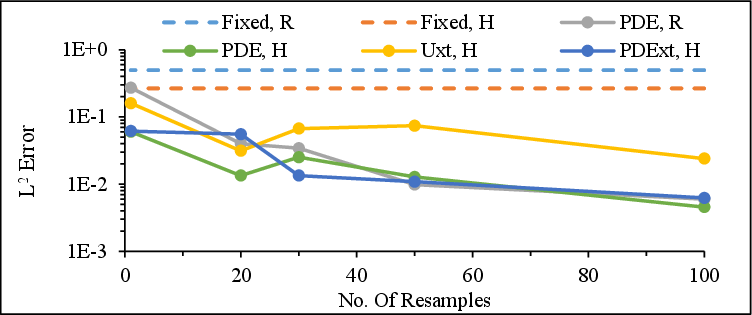}
    \caption{Error against number of resamples solving the Allen-Cahn Equation with $N = 500$}
    \label{fig:ac_n500}
\end{figure}

For $N\geq1000$, a similar pattern is observed to that shown in figure \ref{fig:visc01}, where there is little difference between resampling and fixed methods. In this case, the increased number of points again makes the problem simple enough to solve with fixed collocation strategies, so that the benefits previously observed from using adaptive resampling methods disappear. In fact a fixed pseudo-random Hammersley distribution performs better than the other curves: perhaps again benefiting from more L-BFGS optimiser steps. 

For the Allen-Cahn Equation case, the behaviour and importance of sampling in these methods varies significantly depending on the number of points used. The less sharp shocks in this solution could also be influencing the effectiveness of using derivatives of the solution as redistribution criteria. Nevertheless, the adaptive methods clearly outperform the fixed distribution of collocation points for the case $N=500$, again suggesting that results of a prescribed accuracy can be obtained with fewer collocation points when adaptive sampling is permitted.
\section{Discussion and Conclusion}
\label{sec:5}
In this paper, to better explore the problem of collocation point sampling in PINNs we have examined the two benchmark problems of Burgers' Equation and the Allen-Cahn Equation. We have used novel information sources based on derivatives of the local residual and the solution estimate to guide the resampling process, which has been kept consistent with other approaches in literature, and compared the results to other resampling methods and fixed random and pseudo-random distributions.

Overall, we have observed that using adaptive point distribution typically allows more accurate PINNs solutions to be obtained than using fixed point distributions for small numbers of collocation points. Using fewer collocation points is potentially important since the training cost of a PINN grows in proportion to this number (all other factors being equal). Furthermore, we have observed that defining the probability distribution for the collocation point locations based upon the mixed second derivative of the residual with respect to the independent variables typically outperforms, often significantly, using the residual alone, as in~\cite{wu}.

For Burgers' equation with a smooth initial condition that leads to the formation of a single ``shock'', and a diffusion parameter such that this features is sufficiently sharp, we obtained the smallest errors using this approach (Fig.~\ref{fig:ic1_vs_l}). Equally importantly, we observed that we could reach such errors with significantly fewer collocation points than for any of the other techniques that we compared against (Fig.~\ref{fig:ic1_vs_n}). When considering other initial conditions and other diffusion parameters the relative performance of the adaptive approaches was seen to depend upon the complexity of the specific solution in each case. Typically, there appears to be a value of $N$ above which the use of a uniform (Hammersley) distribution of points is preferred - however this value can be relatively large for the more complex problems (and is relatively small when the solution is smoothly varying everywhere). When considering numbers of points below this critical value of $N$ we invariably find that using the mixed derivative of the residual is the best strategy for defining the PDF for drawing the collocation point locations.

Consideration of a different PDE, the Allen-Cahn equation in this case, demonstrates similar behaviour. Specifically, for a given problem there is a value of $N$ below which the adaptive sampling approach is preferable (and above which a uniform Hammersley distribution suffices). In the particular examples considered here, defining the PDF based upon the mixed derivative of the residual is still reliable and robust, however its improvement relative to using the residual alone is less clear.
Future work will extend the exploration of these resampling methods to other PDEs beyond Allen-Cahn, helping assess the generalizability of our findings. As our approach does not explicitly constrain the type of domain we can study, we should also be able to consider domains with higher spatial dimensions and more complex shapes. 

In terms of computational cost, the PINNs network we have considered is not competitive with standard discretization methods for solving individual forward problems. This is primarily due to the high computational cost of training with full gradient descent methods like L\_BFGS. As for the cost of the resampling strategies we have discussed, these only contribute to a small overhead (estimated at 3\% in \cite{wu}) to the overall computational cost of PINNs training.

Considering other common problems in ML, the biggest challenge is determining what constitutes an appropriate training regime. We did not observe overfitting to occur, even with fixed sampling methods. However, the choice of optimisation algorithm and training regime was still impactful. The use of faster stochastic gradient descent methods like ADAM helped initialise training but had a limited ceiling to accuracy, hence the need for the more expensive L\_BFGS. Whilst we were not able to fully explore the impact of different training regimes within the scope of this paper, efficiency could feasibly be improved by further optimising this via methods such as  cosine annealing \cite{subramanian}; or by optimising the architecture employed for solving problems of varying complexity.

In summary, we have shown that a number of different strategies are viable for selecting the location of PINNs collocation points with resampling methods. In particular, we have shown that resampling can be guided not only by the local residuals, but also by the spatial and temporal derivatives of the residuals and even of the solution estimates themselves. Furthermore, we have observed that a number of different factors impact the accuracy of the PINN method, making it difficult to analyse any single aspect (such as collocation point selection) in isolation. Whilst we believe that our contribution begins to shed some light on the interplay between the location of the collocation points and the ability of a PINN to learn the PDE solution it is clear that further research is desirable in order to be able to propose robust implementations that support reliable training or provide accuracy guarantees.

\begin{credits}
\subsubsection{\ackname} 
The first author gratefully acknowledges financial support via the EPSRC Centre for Doctoral Training in Fluid Dynamics at Leeds (EP/S022732/1).

\subsubsection{\discintname}
The authors have no competing interests to declare that are
relevant to the content of this article.

\end{credits}

\end{document}